\documentclass{ecai} 

\usepackage{latexsym}
\usepackage{amssymb}
\usepackage{amsmath}
\usepackage{amsthm}
\usepackage{booktabs}
\usepackage{enumitem}
\usepackage{graphicx}
\usepackage{color}
\usepackage{bm}
\usepackage{array}
\usepackage{pifont}
\usepackage{xcolor}
\usepackage{multirow}
\usepackage{subfigure}

\newcommand{\BibTeX}{B\kern-.05em{\sc i\kern-.025em b}\kern-.08em\TeX}

\begin{document}


\begin{frontmatter}


\paperid{5400} 


\title{Multi-Modal Continual Learning via \\ Cross-Modality Adapters and Representation \\Alignment with Knowledge Preservation}


\author[A]{\fnms{Evelyn}~\snm{Chee}\thanks{Corresponding Author. Email: evelyn.chee@u.nus.edu}}
\author[A]{\fnms{Wynne}~\snm{Hsu}}
\author[A]{\fnms{Mong Li}~\snm{Lee}} 

\address[A]{School of Computing, National University of Singapore}


\begin{abstract}
Continual learning is essential for adapting models to new tasks while retaining previously acquired knowledge. 
While existing approaches predominantly focus on uni-modal data, multi-modal learning offers substantial benefits by utilizing diverse sensory inputs, akin to human perception. However, multi-modal continual learning presents additional challenges, as the model must effectively integrate new information from various modalities while preventing catastrophic forgetting. 
In this work, we propose a pre-trained model-based framework for multi-modal continual learning. 
Our framework includes a novel cross-modality adapter with a mixture-of-experts structure to facilitate effective integration of multi-modal information across tasks. 
We also introduce a representation alignment loss that fosters learning of robust multi-modal representations, and regularize relationships between learned representations to preserve knowledge from previous tasks.
Experiments on several multi-modal datasets demonstrate that our approach consistently outperforms baselines in both class-incremental and domain-incremental learning, achieving higher accuracy and reduced forgetting.
\end{abstract}

\end{frontmatter}


\section{Introduction}

Continual learning in neural networks has received significant attention due to its relevance in real-world applications, where models need to continuously adapt to new distributions or learn new classes. The primary challenge in continual learning is to retain previously learned knowledge as new knowledge are acquired. 
To address this forgetting issue, existing approaches can be broadly categorized into: rehearsal-based techniques \cite{aljundi2019gradient,chaudhry2019efficient,rebuffi2017icarl} that store a small number of past samples for subsequent learning; regularization-based techniques \cite{buzzega2020dark,douillard2020podnet,kang2022class} that penalize divergence from previous models; and architecture-based techniques \cite{chee2023leveraging,hung2019compacting,mallya2018piggyback,wang2022foster} that dynamically expand the model to accommodate new knowledge. 

Advancements in parameter efficient fine-tuning \cite{chen2022adaptformer,houlsby2019parameter} and the availability of large-scale pre-trained models (PTMs)  have led to new approaches that leverage these models in continual learning. These PTM-based continual learning approaches focus on learning a small number of tunable parameters, such as prompt tuning \cite{smith2023coda,wang2022dualprompt,wang2022learning} and adapter tuning \cite{tan2024semantically,yu2024boosting,zhou2024expandable}. 
While these approaches have shown promise, they have primarily been applied to  single-modality data.

Deep learning models that utilize multi-modal data have shown significant advantage over their uni-modal counterparts \cite{jabeen2023review,ramachandram2017deep}, akin to how human sensory systems integrate diverse inputs to enhance perception and improve predictive accuracy.
This motivates the study of multi-modal continual learning, which aims to seamlessly integrate information from multiple modalities while retaining previously acquired knowledge \cite{mo2023class,pian2023audio,wang2023confusion}.

\begin{figure}[t]
    \centering
    \includegraphics[width=0.98\linewidth]{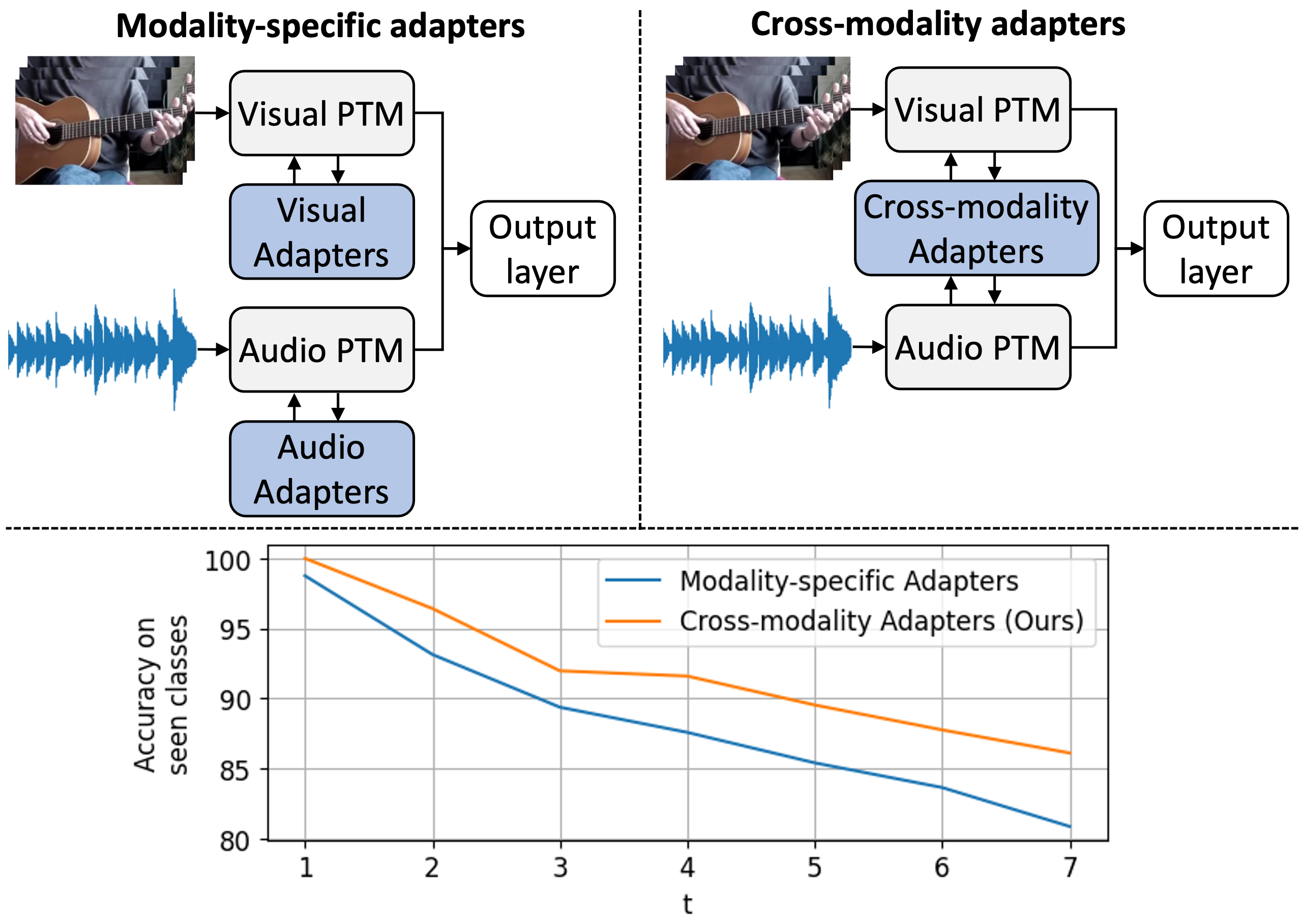}
    \caption{ Modality-specific adapters versus our proposed cross-modality adapters in continual learning. Results on  AVE dataset \cite{tian2018audio} show that learning modality interactions enhances accuracy on seen classes. } 
    \label{fig:intro}
\end{figure}

A direct approach to incorporating PTMs into multi-modal continual learning is to utilize models independently pre-trained on each input modality. However, this approach overlooks the interactions between modalities, limiting the model's ability to adapt to new tasks and compromising performance in multi-modal settings.
To address this challenge, we propose a PTM-based framework for multi-modal continual learning that effectively integrates cross-modality interactions.
Our approach introduces cross-modality adapters that capture interactions across embeddings from corresponding layers of modality-specific PTMs to facilitate adaptation of the PTM features. 

We design the cross-modality adapters with a mixture-of-experts (MoE) structure \cite{shazeer2017outrageous}, comprising expert networks that continuously learn cross-modality interactions for newly introduced tasks and a gating network that selects the most suitable experts for processing each task. 
To prevent forgetting of learned knowledge, we preserve the integrity of experts that were actively used in earlier tasks by freezing their weights.
In contrast to using separate adapters for each modality, our approach enables effective integration of multi-modal information while maintaining the flexibility to incorporate new knowledge in a continual learning setting (see Figure \ref{fig:intro}).

Additionally, we introduce a representation alignment loss to learn robust representations for the final classification. Traditional multi-modal contrastive learning frameworks \cite{li2021align,radford2021learning} have shown effectiveness by aligning representations of all modalities into a common space. However, our experiments show that directly using the contrastive loss leads to suboptimal performance. This is mainly due to its strict cross-modality alignment constraint, which limits the model's ability to capture modality-specific features  \cite{dong2024simmmdg}. 
The restricted set of features can cause overlap between representations, such as those of old and new classes, making them harder to distinguish and negatively impacting overall classification performance. 

Instead of directly enforcing alignment between modalities, our proposed loss aligns each modality’s representation independently with a joint representation derived from fusing the different modalities. Since the joint representation captures information from all modalities, aligning each modality with it reduces information loss and results in more stable representations. 
To further preserve knowledge, we apply a regularization term that penalizes changes in the relationships between previously learned representations.

To evaluate our approach, we conduct experiments on various multi-modal classification datasets with diverse modality combinations, across different learning scenarios such as class-incremental and domain-incremental learning. 
We compare our method against state-of-the-art continual learning approaches, including recent PTM-based approaches.
Empirical results show that our method yields significant performance improvements across these scenarios, achieving both higher overall accuracy and reduced forgetting.

\section{Related Work}

Existing continual learning methods largely focus on mitigating forgetting.
Rehearsal-based approaches address this by storing a subset of past samples and replaying them to reinforce prior knowledge.
For example, Experience Replay (ER) \cite{chaudhry2019efficient} trains on a mix of new and stored samples at each phase. 
iCaRL \cite{rebuffi2017icarl} improves this by  selecting and storing  more representative samples.
GDumb \cite{prabhu2020gdumb} defers training entirely until inference, using only stored samples.
Regularization-based methods  introduce loss terms that constrain the updated model from diverging significantly from previous states. 
For instance, PODNet \cite{douillard2020podnet} enforces consistency in intermediate features, while DER++ \cite{buzzega2020dark} uses distillation loss to preserve past predictions.
Architecture-based solutions, such as FOSTER \cite{wang2022foster} and MEMO \cite{zhou2023model}, incrementally expand the model with new parameters for incoming tasks, while freezing earlier ones to preserve prior knowledge.

The availability of large PTMs has led to recent continual learning methods that adapt these models to new tasks using a small set of additional trainable parameters.
Prompt-based methods, such as L2P \cite{wang2022learning}, DualPrompt \cite{wang2022dualprompt} and CODA-Prompt \cite{smith2023coda}, learn a pool of prompts to condition the PTM for each new task and select appropriate prompts at inference time to guide prediction.
Adapter-based methods, including  APER \cite{zhou2024revisiting}, EASE \cite{zhou2024expandable} and InfLoRA \cite{liang2024inflora}, insert  lightweight adapter layers into the frozen PTM and train the adapters to learn task-specific knowledge.

These prior works have focused on uni-modal tasks, particularly with imaging datasets.
In multi-modal continual learning \cite{yu2024recent}, recent studies \cite{mo2023class,pian2023audio,wang2023confusion,wang2022s,yu2024boosting} address the challenges of learning from multiple modalities over time.
For example, AV-CIL \cite{mo2023class} uses distillation on audio-guided visual attention maps to preserve relevant features, while CMR-MFN \cite{wang2023confusion} adopts an expandable architecture to model evolving visual-inertial interactions.
However, these methods remain tailored to specific modality pairs, limiting their applicability to broader multi-modal scenarios.

Our work is also closely related to cross-modal representation alignment, a key component in multi-modal learning. Various works \cite{du2024delan,jia2021scaling,pandey2023cross,radford2021learning} employ objectives such as contrastive loss and alignment loss to project different modalities into a shared space, enabling effective cross-modal interaction and improving model performance. However, these methods assume access to the full data distribution, including all classes, throughout training and can in fact worsen forgetting when applied in continual learning scenarios.

In summary, this work bridges continual learning, multi-modal learning, and cross-modal representation alignment by extending adapter-based PTM methods to flexible multi-modal settings and employing mechanisms for robust representation learning over time.

\begin{figure}[t]
    \centering
    \includegraphics[width=\linewidth]{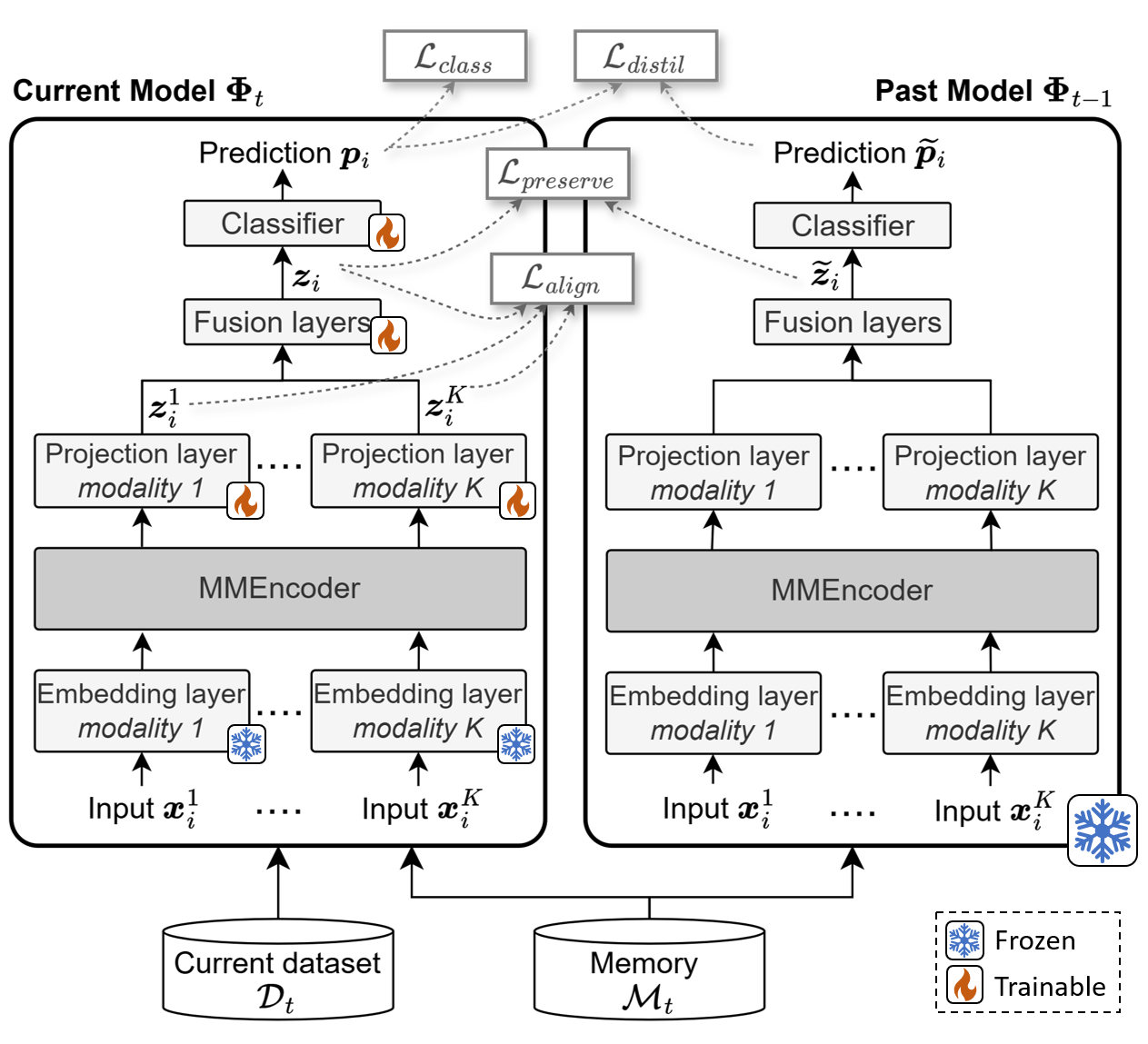}
    \caption{Proposed multi-modal continual learning framework at time $t$. }
    \label{fig:overview}
\end{figure}

\section{Proposed Method}

Multi-modal continual learning aims to develop a model through a sequence of $T$ tasks, each involving samples with $K>1$ input modalities. At each time step, a new task is introduced with limited access to data from previous tasks. 

We consider two continual learning scenarios: class-incremental learning and domain-incremental learning. In class-incremental learning, new classes are introduced at each time step, with no overlap between the new and previously seen classes. Formally, if $\mathcal{C}_t$ denotes the set of classes at time step $t$,  then $\mathcal{C}_i\cap\mathcal{C}_j=\emptyset$ for all $i\neq j$.
For domain-incremental learning, the set of classes remains fixed, {\it i.e.} $\mathcal{C}_i=\mathcal{C}_j$ for all $i,j\in[1,T]$, 
but the input data distribution shifts across tasks.
In both scenarios, the model is required to minimize prediction error across all tasks learned.

Figure \ref{fig:overview} provides an overview of our proposed framework for multi-modal continual learning. 
At each time step $t$, we update the previously learned model $\Phi_{t-1}$ using the combined dataset $\mathcal{D}_t\cup\mathcal{M}_t$ to obtain an updated model $\Phi_{t}$. Here, the dataset $\mathcal{D}_t=\{(\bm{x}_i,\bm{y}_i)\}$ represents newly arriving data at time $t$, where $\bm{x}_{i} = \{\bm{x}^1_i, \cdots \bm{x}_i^ K\}$ consists of $K$ different modalities for the $i^{th}$ data instance, and $\bm{y}_i$ denotes its corresponding one-hot encoded class label. Meanwhile, the set $\mathcal{M}_t$ refers to the retained data instances from previous time steps, which are used for replay.

Our model is built upon transformer-based PTMs specifically trained for each modality $k\in[1,K]$. 
We freeze the weights of these PTMs during training.
Each input modality $\bm{x}_{i}^k$ is first tokenized using the corresponding PTM's input embedding layer.
The tokenized inputs are forwarded to our Multi-Modal Encoder (MMEncoder), which captures the interaction between modalities and outputs a representation for each modality.
These representations are then processed by projection layers to obtain modality-specific representations $\bm{z}^k_i$, all of which have the same dimension.
The modality-specific representations are concatenated and passed through fusion layers to generate joint representation $\bm{z}_i$.
This joint representation is then used by the classifier to output prediction $\bm{p}_i$.

To enhance the robustness of the learned multi-modal representations and mitigate forgetting, we introduce two loss functions: 
$\mathcal{L}_{align}$ for representation alignment and $\mathcal{L}_{preserve}$ for minimizing the divergence between the current representations and those learned in the previous time step.

The following subsections provide detailed descriptions of our proposed components: the MMEncoder in Section \ref{sec:mmencoder} and the loss functions in Section \ref{sec:loss}.

\begin{figure}[t!]
    \centering
    \includegraphics[width=0.99\linewidth]{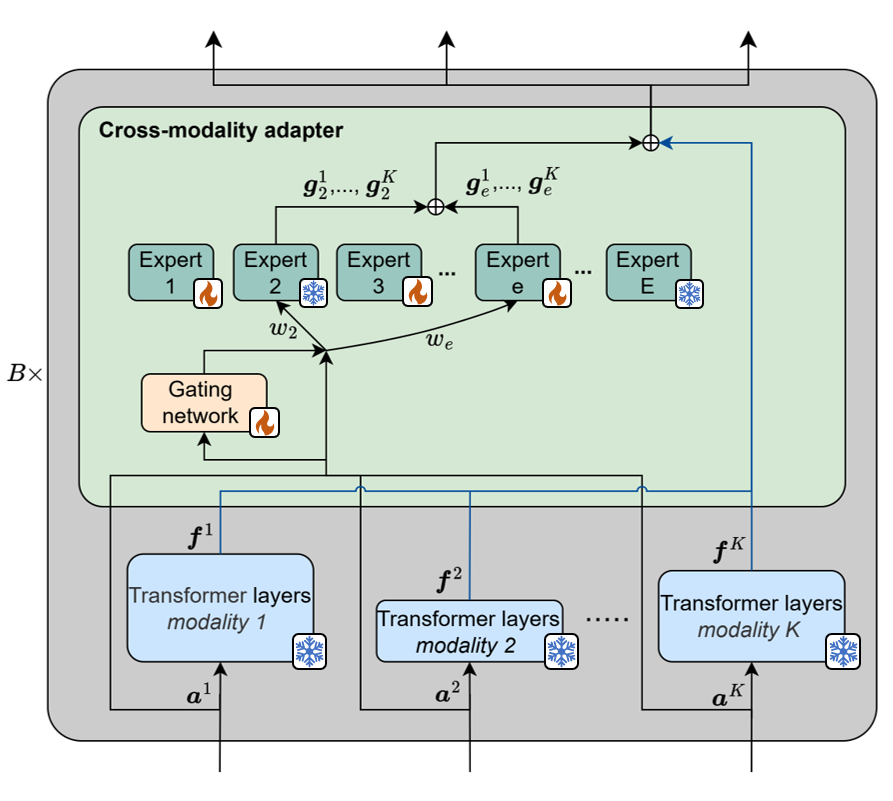}
    \caption{MMEncoder with a cross-modality adapter in each block.}
    \label{fig:cross_modality}
\end{figure}

\subsection{MMEncoder}
\label{sec:mmencoder}
The key cornerstone in our framework is the MMEncoder, which adapts features from the transformer layers of the modality-specific PTMs by incorporating our novel cross-modality adapters. These adapters are designed to capture the interactions between the various modalities.

Our MMEncoder is composed of $B$ blocks, where each block contains a number of consecutive transformer layers from the PTMs of all modalities. 
Since each PTM may have different number of transformer layers, denoted as $L^k$ for modality $k$, we set $B= \min_{k = 1}^K L^k$ and divide the layers of each modality's PTM into $B$ blocks.
Specifically, the number of transformer layers for modality $k$ assigned to block $b\in[1,B]$ is determined as follows:
\begin{equation}
\begin{cases}
  1 + \lfloor \frac{L^k}{B} \rfloor & \text{if } b \leq L^k - B\lfloor \frac{L^k}{B} \rfloor, \\
  \lfloor \frac{L^k}{B} \rfloor & \text{otherwise.}
\end{cases}    
\end{equation}
This formulation ensures that the transformer layers are distributed as evenly as possible across all the blocks.
Any remainder from dividing $L^k$ by $B$ is handled by allocating one additional layer to each of the first $L^k - B \left\lfloor \frac{L^k}{B} \right\rfloor$ blocks.

Within each block, we add a cross-modality adapter as shown in Figure \ref{fig:cross_modality}.  The cross-modality adapter employs a mixture-of-expert structure, consisting of $E$ experts and a gating network. 
This design allows each expert to specialize in specific  cross-modality interactions learned over time, while the gating mechanism ensures that only the relevant experts are activated for a given input, offering both flexibility and scalability required for continual learning.

The adapter takes as input both $\bm{a}^k$ and $\bm{f}^k$ for all $k \in [1..K]$, where $\bm{a}^k$ is the input for modality $k$ to the block, and $\bm{f}^k$ is the output after processing by the transformer layers of the corresponding modality in the block.
To determine which experts to activate for a given input, the gating network is trained to use $\bm{a}^1, \dots, \bm{a}^K$ to generate the contribution weights $w_{1}, \dots, w_{E}$ for each of the $E$ experts:
\begin{equation}
[w_{1}, \dots, w_{E}] = \mathcal{G}(\bar{\bm{a}}^1 \oplus \cdots \oplus~\bar{\bm{a}}^K), 
\end{equation}
where $\bar{\bm{a}}^k$ 
denotes the average of $\bm{a}^k$ across the sequence dimension, and $\oplus$ denotes the concatenation operation.
$\mathcal{G}(\cdot)$ projects the concatenated averages into a 1-dimensional vector, where each entry represents the activation likelihood for a corresponding expert. 
These likelihood are normalized into weights within the range of $[0,1]$ using the softmax function.
The resulting weights signify the relative competencies of the experts for the given input. A subset of the experts with the highest non-zero weights are then activated.

Each activated expert $e$  accounts for the cross-modality interactions by first
downsizing the inputs $\bm{a}^k$ from each modality
$k$  
to the same feature dimension using a down-projection layer $\bm{W}^{k}_{down,e}$, followed by a ReLU activation to obtain $\bm{h}_e^k$:
\begin{equation}
\bm{h}_e^k = \text{ReLU}(\bm{a}^k\bm{W}^{k}_{down,e}). 
\end{equation}
For each modality $k$, we then generate an attention map  $\bm{m}_e^k$, which captures information from the other modalities. 
This is achieved by  averaging $\bm{h}_e^j$ from the other modalities, $j \neq k$, across the sequence dimension, and summing them with  $\bm{h}_e^k$.
The resulting features are projected back to the respective original dimensionality using an attention layer $\bm{W}^{k}_{att,e}$ and subsequently passed through a tanh activation function. Specifically, we have:
\begin{equation}
\bm{m}_e^k = \text{tanh}\Big((\bm{h}_e^k+\sum_{j\neq k}\bar{\bm{h}}_e^j)\bm{W}^{k}_{att,e}\Big),
\end{equation}
where $\bar{\bm{h}}_e^j$ denotes the average of  $\bm{h}_e^j$ across the sequence dimension.

Next, we incorporate the attention map information by upsizing  $\bm{h}_e^k$ using an up-projection layer $\bm{W}^{k}_{up,e}$, followed by element-wise multiplication with the attention map $\bm{m}_e^k$:
\begin{equation}
\bm{g}_e^k = (\bm{h}_e^k\bm{W}^{k}_{up,e}) \odot \bm{m}_e^k.
\end{equation}
This enables each expert to modulate the features of each modality by leveraging the relevant information from other modalities as captured by the attention map.
The output of the experts  $\bm{g}_e^k$ are used to adapt the output $\bm{f}^k$ from the PTM transformer layers of the corresponding modality within the block as follows: 
\begin{equation}
    \bm{f}^k + \sum_{e}w_{e}\bm{g}^k_{e},
\end{equation}
where $w_{e}$ is the weight of the expert $e$. 
The resulting  adapted features are then passed as the input $\bm{a}^k$ to the next block.

\begin{figure}[t!]
    \centering
    \includegraphics[width=0.99\linewidth]{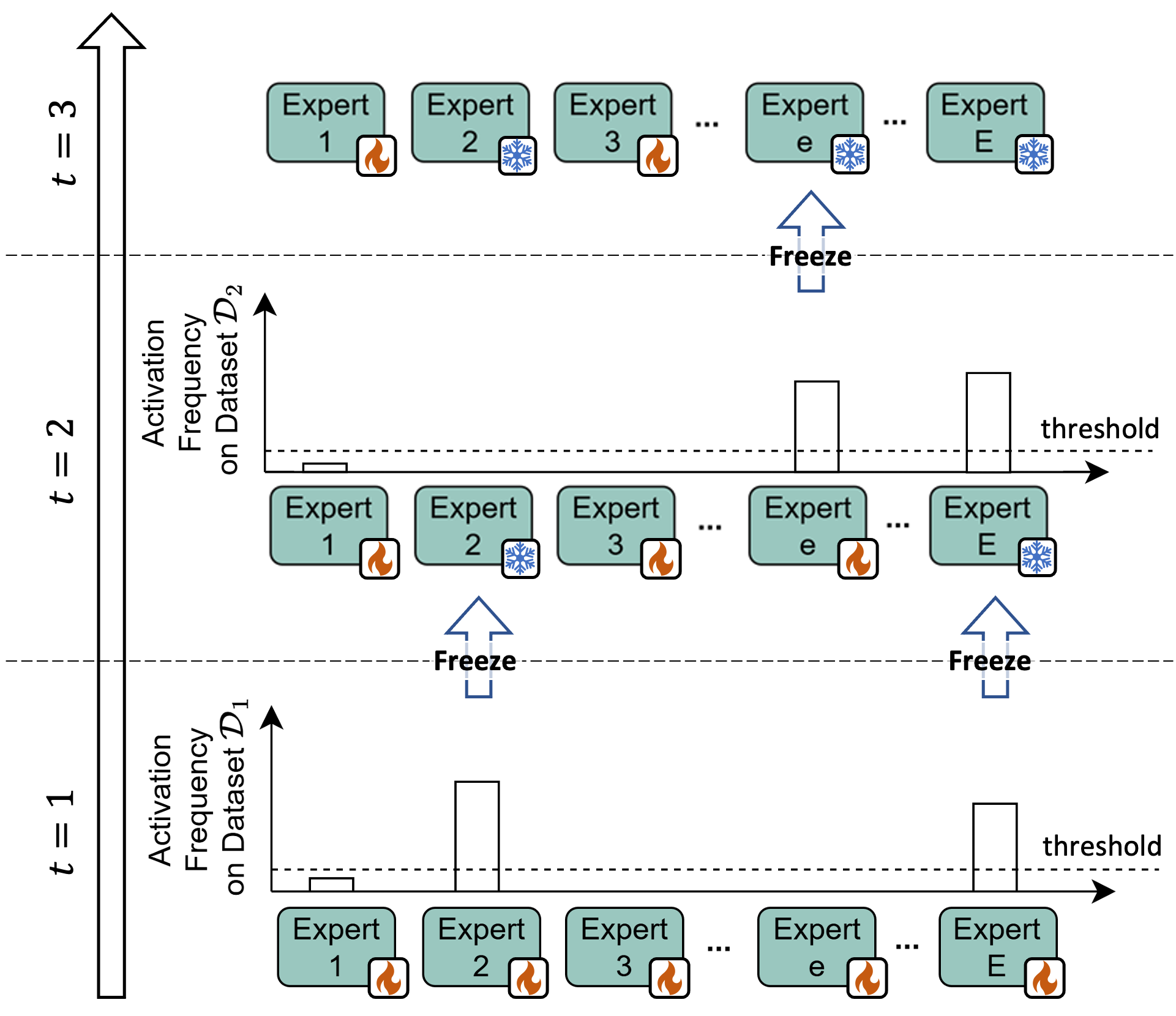}
    \caption{Trainable experts as learning progresses across time steps.}
    \label{fig:learning_expert}
\end{figure}

To preserve the knowledge of cross-modality interactions learned across tasks, we freeze the weights of the experts.
In particular, after training of each task at time step $t$, we perform a single-pass inference on the current dataset $\mathcal{D}_t$ to compute the activation frequency of each expert. 
As illustrated in Figure  \ref{fig:learning_expert}, experts activated more than a predetermined threshold, which we set as a fixed percentage of the total samples in $\mathcal{D}_t$, are kept frozen in subsequent time steps.
When learning a new task, the model can  leverage these frozen experts to utilize prior knowledge or use the trainable experts to incorporate new information. 

\subsection{Representation Alignment}
\label{sec:loss}
Our goal is for the MMEncoder to learn multi-modal representations that remain robust as learning progresses, ensuring that the model effectively adapts to new knowledge without forgetting previously acquired information. To achieve this, we first focus on preserving the distinct information inherent in each modality while maintaining consistent alignment across the different modalities. 

\begin{figure}[t!]
    \centering
    \includegraphics[width=0.97\linewidth]{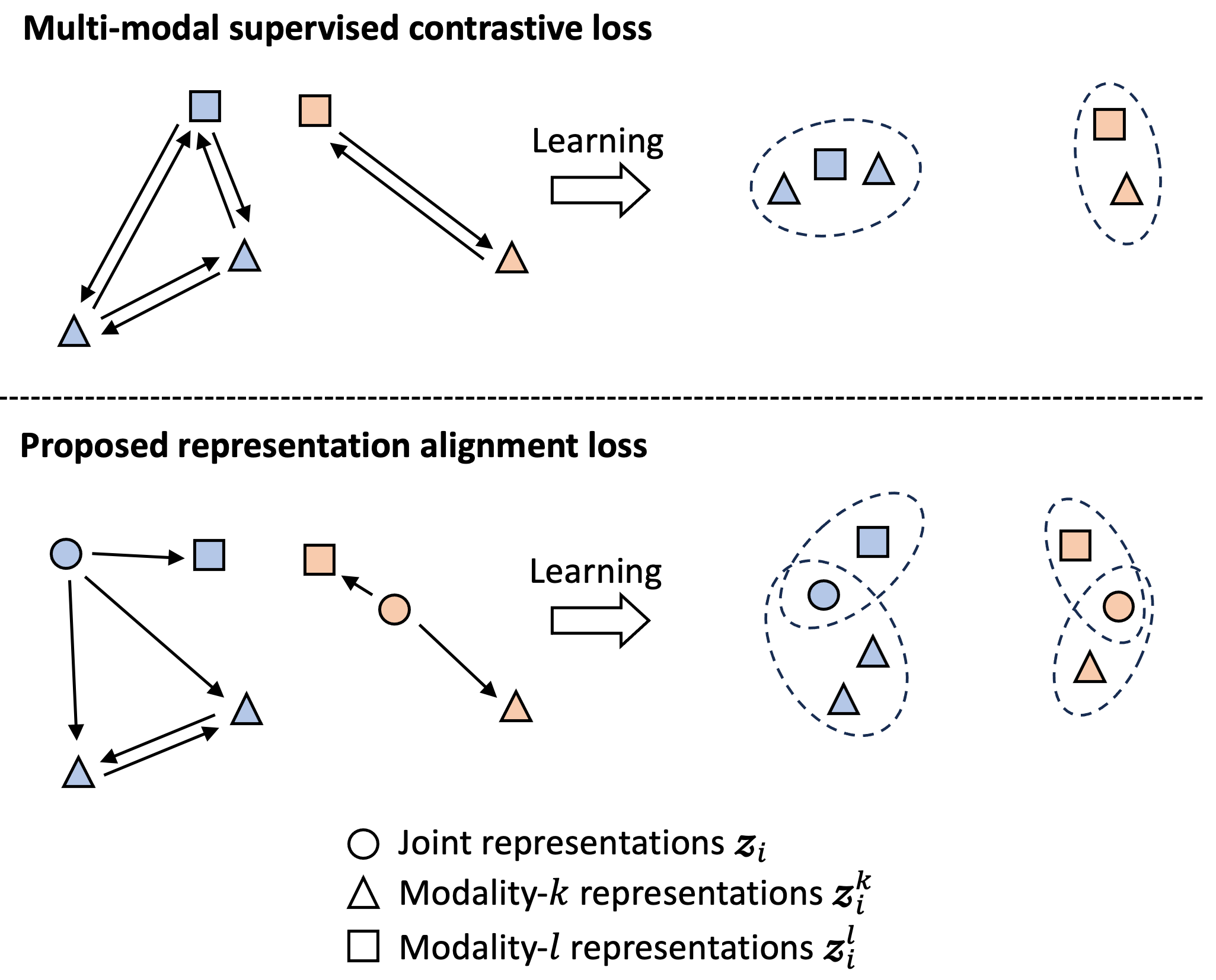}
    \caption{Conventional multi-modal supervised contrastive loss versus our proposed representation alignment loss, where modality-specific representations are aligned independently. Symbols with the same color indicate representations of the same class.}
    \label{fig:loss}
\end{figure}

A common approach to align representations of different modalities is to map them into a shared feature space using contrastive learning \cite{li2021align,radford2021learning}. However, this often leads to loss of modality-specific information \cite{dong2024simmmdg}.
As illustrated in Figure \ref{fig:loss}, the strict constrain imposed by supervised contrastive loss to directly align representations of different modalities can result in clusters containing mixed modalities, suggesting a loss of modality-specific information.

To address this issue, we propose aligning each modality-specific representation independently with the joint representation, rather than enforcing direct cross-modality alignment.
Since the joint representations encompasses information from all modalities, this indirectly leads to alignment of all modalities into a shared space. 
We preserve modality-specific information by aiming for each modality to form distinct clusters, while still keeping the different modalities of the same class close together.

Recall that $\bm{z}_{i}^{k}$ denotes the representation obtained by projecting the output of modality $k$ from the MMEncoder for sample $\bm{x}_i$, and $\bm{z}_i$ is the joint representation obtained by projecting the concatenated representations $\bm{z}_{i}^{1} \oplus \cdots \oplus \bm{z}_{i}^{K}$ to the same dimension as each $\bm{z}_{i}^{k}$.
Using $\bm{z}^k_{i}$ as anchor, we want to pull representations of samples from the same class closer to the anchor, while pushing those from different classes away.

Consider a batch of $N$ samples drawn from the dataset $\mathcal{D}_t\cup\mathcal{M}_t$.
Let $\mathcal{J}=\{\bm{z}_j \mid  j\in [1,N]\}$ represent the set of joint representations for these samples, and $\mathcal{S}_i^k=\{\bm{z}_{j}^{k} \mid i\neq j, j \in [1,N]\}$ represent the set of modality-specific representations for modality $k$, excluding the anchor $\bm{z}_{i}^{k}$. 
We define $\mathcal{U}^k_i=\mathcal{S}_i^k\cup\mathcal{J}$ as the union of these two representation sets, and $\mathcal{V}^k_i\subset\mathcal{U}^k_i$ as the subset of representations corresponding to samples with the same class as $\bm{x}_i$.

Using representations from the same modality, along with the joint representations, we define the alignment loss for an anchor $\bm{z}_i^k$ as:  
\begin{equation}
    \label{eq:ccur_k}
    \text{align}(\bm{z}_i^k)= \frac{-1}{|\mathcal{V}^k_{i}|} \sum_{\bm{v}\in \mathcal{V}_{i}^{k}}\log \frac{\exp (\bm{z}_{i}^{k}\boldsymbol{\cdot}\bm{v}/\tau)}{\sum_{\bm{u}\in \mathcal{U}_{i}^{k}}\exp (\bm{z}_{i}^{k}\boldsymbol{\cdot}\bm{u}/\tau)}, 
\end{equation}
where $\boldsymbol{\cdot}$ denotes the cosine similarity between two normalized representations, and $\tau$ is a temperature hyperparameter.
Our overall representation alignment loss is then computed as follows:
\begin{equation}
    \label{eq:ccur}
    \mathcal{L}_{align} = \frac{1}{K|\mathcal{I}_\mathcal{D}|}\sum_{k\in[1,K]} \sum_{i\in\mathcal{I}_\mathcal{D}} \text{align}(\bm{z}_i^k),
\end{equation}
where $\mathcal{I}_\mathcal{D}\subset \{1,\dots,N\}$ is the set of indices for samples in the batch that belong to the current dataset $\mathcal{D}_t$.
We only use representations from current task data as anchors because the data from previous tasks is limited. Optimizing the loss on these few instances could lead to overfitting of previous task knowledge, thereby accelerating forgetting.

While $\mathcal{L}_{align}$ helps reduce loss of modality-specific information when learning new tasks, it does not explicitly mitigate forgetting of previously learned knowledge. 
As such,
we impose a regularization loss that ensures consistency between the learned representations as learning progresses.

Let $\tilde{\bm{z}}_{i}$ and $\bm{z}_{i}$ be the joint representations of sample $\bm{x}_i$, extracted using the model from the previous time step, $\Phi_{t-1}$, and the current model, $\Phi_t$, respectively.
We quantify the relation between pairs of memory samples by computing the normalized pairwise similarity between their joint representations from each model as follows:
\begin{align}    
\begin{split}
 \widetilde{s}_{i,j} &= \frac{\exp(\widetilde{\bm{z}}_{i} \boldsymbol{\cdot} \widetilde{\bm{z}}_{j}/\tau)}{\sum_{m\in {\mathcal{I}_\mathcal{M}}\setminus\{i\}}\exp(\widetilde{\bm{z}}_{i} \boldsymbol{\cdot}, \widetilde{\bm{z}}_{m}/\tau)}, \\
 s_{i,j} &= \frac{\exp(\bm{z}_{i} \boldsymbol{\cdot} \bm{z}_{j}/\tau)}{\sum_{m\in \mathcal{I}_\mathcal{M}\setminus\{i\}}\exp(\bm{z}_{i} \boldsymbol{\cdot} \bm{z}_{m}/\tau)},   
\end{split}
\end{align}    
where $i,j \in \mathcal{I}_\mathcal{M}$ and $\mathcal{I}_\mathcal{M} \subset \{1,\dots,N\}$ is the set of indices for memory samples in the batch. 

To preserve the relations between memory samples, we minimize the Kullback–Leibler divergence between the two sets of pairwise similarity scores:
\begin{equation}
    \label{eq:cpre}
    \mathcal{L}_{preserve} = \frac{\sqrt{t-1}}{|{\mathcal{I}_\mathcal{M}}|}\sum_{i\in {\mathcal{I}_\mathcal{M}}}\sum_{j\in {\mathcal{I}_\mathcal{M}}\setminus\{i\}}\widetilde{s}_{i,j}\log\frac{\widetilde{s}_{i,j}}{s_{i,j}}.
\end{equation}
We weight the loss to account for the increasing importance of preserving previously learned knowledge as more tasks are learned. Note that $\mathcal{L}_{preserve}$ is not applied during the first time step $t=1$.
 
\subsection{Overall Optimization Objective}

In addition to our representation learning losses, we include two other losses commonly used in continual learning to optimize the classifier weights: classification loss and logits-distillation loss. 
For classification loss, we apply the standard cross-entropy loss to distinguish concepts across all tasks seen so far:
\begin{equation}
\mathcal{L}_{class}=\frac{1}{N}\sum_{i\in [1,N]}\sum_{c\in \mathcal{C}_{1:t}} y_{i,c}\log p_{i,c} ~,
\end{equation}
where $\mathcal{C}_{1:t}$ is the set of all classes seen up to time step $t$, $y_{i,c}$ is the ground truth of sample $\bm{x}_i$ for class $c$ and $p_{i,c}$ is the corresponding predicted probability, obtained from logits $\bm{p}_i$ using the softmax function for single-class classification tasks and the sigmoid function for multi-class classification tasks.

To prevent information loss at the decision-level, the logits-distillation loss is applied to the predicted logits of memory samples \cite{buzzega2020dark} as follows: 
\begin{equation}
    \label{eq:dis}
    \mathcal{L}_{distil} = \frac{\sqrt{t-1}}{|\mathcal{I}_\mathcal{M}|}\sum_{i\in {\mathcal{I}_\mathcal{M}}}\lVert \bm{p}_{i} - \widetilde{\bm{p}}_{i} \rVert_2^2~,
\end{equation}
where $\widetilde{\bm{p}}_{i}$ and $\bm{p}_{i}$ are the logits of sample $\bm{x}_i$ output by the past model $\Phi_{t-1}$ and the current model $\Phi_t$ respectively. Similar to $\mathcal{L}_{preserve}$, this loss is not applied at $t=1$ and its weight increases as $t$ increases.

Finally, the overall loss function combines the classification loss, distillation loss, and our proposed representation alignment with preservation loss:
\begin{equation}
    \mathcal{L} = \mathcal{L}_{class} + \lambda_1\mathcal{L}_{distil} + \lambda_2\mathcal{L}_{align} + \lambda_3\mathcal{L}_{preserve}, 
\end{equation}
where $\lambda_1$, $\lambda_2$ and $\lambda_3$ are hyperparameters for balancing the losses.
The weights of the trainable experts, gating networks, projection layers, fusion layers, and classifier layer are optimized using this combined loss.

\section{Performance Study}
\paragraph{Dataset and Learning Setup.} 
Our evaluation covers both class-incremental learning and domain-incremental learning scenarios. For class-incremental learning, we  use the AVE and UESTC-MMEA datasets. For domain-incremental learning, we utilize the SAMSEMO dataset.
\begin{itemize}
\item AVE \cite{tian2018audio}: This audio-visual dataset consists of 28 event classes, such as human activities, animal behaviors, music instrument performance and vehicle sounds. For incremental learning, we adapt the dataset by randomly splitting the classes into seven distinct sets, with each containing 4 classes that are introduced at different time steps. 
\item UESTC-MMEA \cite{xu2023towards}: This egocentric dataset utilizes visual and inertial data, covering 32 activity classes that range from static activities like reading and watching television, to physical activities such as climbing stairs and cycling. 
For incremental learning, we randomly split the classes into eight disjoint sets, each consisting of 4 classes introduced across different time steps. 
\item SAMSEMO \cite{bujnowski2024samsemo}: This is a multilingual emotion recognition dataset with visual, audio and text modalities. It consists scenes in five languages -- English, German, Spanish, Polish and Korean -- each labeled with up to two emotions: anger, happiness, sadness, surprise and neutral. At each time step, data from a new language is introduced as the new domain.   
\end{itemize}
For all experiments, we limit the memory to a maximum of 200 samples at every time step. Each experiment is repeated across three different class or domain order sequences, and we report the average performance.

\begin{table*}[t!]
\caption{Results of comparative study. }
\label{tab:comparative_results}
    \centering
    \begin{tabular}{>{\raggedright\arraybackslash}m{2.5cm}|>{\centering\arraybackslash}m{1.7cm}>{\centering\arraybackslash}m{1.7cm}|>{\centering\arraybackslash}m{1.7cm}>{\centering\arraybackslash}m{1.7cm}|>{\centering\arraybackslash}m{1.7cm}>{\centering\arraybackslash}m{1.7cm}}
    \toprule
    & \multicolumn{2}{c|}{AVE} & \multicolumn{2}{c|}{UESTC-MMEA} & \multicolumn{2}{c}{SAMSEMO} \\
    \cmidrule{2-7}
    Method & $Acc$ & $Fgt$ & $Acc$ & $Fgt$ & $Acc$ & $Fgt$ \\
     \midrule
    ER & 80.06{ \scriptsize $\pm$1.52} & 13.72{ \scriptsize $\pm$1.43} & 88.73{ \scriptsize $\pm$0.78} & 8.06{ \scriptsize $\pm$0.68} & 53.84{ \scriptsize $\pm$6.63} & 11.01{ \scriptsize $\pm$5.76} \\
    iCaRL & 82.50{ \scriptsize $\pm$3.23} & 12.73{ \scriptsize $\pm$2.87} & 90.55{ \scriptsize $\pm$0.84} & 6.26{ \scriptsize $\pm$0.87} & 53.78{ \scriptsize $\pm$2.60} & 4.04{ \scriptsize $\pm$1.58} \\
    DER++ & 81.60{ \scriptsize $\pm$2.22} & 11.05{ \scriptsize $\pm$1.42}  & 90.02{ \scriptsize $\pm$0.26}  & 6.53{ \scriptsize $\pm$0.12}  & 44.48{ \scriptsize $\pm$8.82}  & 18.22{ \scriptsize $\pm$8.01} \\
    GDumb & 80.82{ \scriptsize $\pm$1.79} & 6.07{ \scriptsize $\pm$0.78}  & 90.52{ \scriptsize $\pm$0.99}  & 3.67{ \scriptsize $\pm$0.26}  & 51.00{ \scriptsize $\pm$1.89}  & 5.38{ \scriptsize $\pm$0.76}  \\
    L2P & 82.40{ \scriptsize $\pm$1.21} & 13.20{ \scriptsize $\pm$1.65} & 88.35{ \scriptsize $\pm$1.37} & 8.06{ \scriptsize $\pm$1.02} & 55.25{ \scriptsize $\pm$0.85} & 13.52{ \scriptsize $\pm$3.17} \\
    DualPrompt & 82.79{ \scriptsize $\pm$1.40} & 11.92{ \scriptsize $\pm$0.70} & 88.11{ \scriptsize $\pm$1.32} & 8.45{ \scriptsize $\pm$1.16} & 55.72{ \scriptsize $\pm$0.42} & 13.18{ \scriptsize $\pm$3.23} \\
    CODA-Prompt & 82.04{ \scriptsize $\pm$0.76} & 13.05{ \scriptsize $\pm$1.09} & 89.12{ \scriptsize $\pm$1.79} & 7.33{ \scriptsize $\pm$1.64} & 55.21{ \scriptsize $\pm$1.07} & 11.76{ \scriptsize $\pm$3.74}  \\
    APER w/ Adapter & 81.99{ \scriptsize $\pm$0.52} & 3.69{ \scriptsize $\pm$0.99} & 89.42{ \scriptsize $\pm$1.01} & 1.40{ \scriptsize $\pm$0.41} & - & - \\
    APER w/ SSF & 83.05{ \scriptsize $\pm$1.48} & 4.72{ \scriptsize $\pm$0.46} & 90.10{ \scriptsize $\pm$1.48} & 1.37{ \scriptsize $\pm$0.56} & - & - \\
    EASE & 82.68{ \scriptsize $\pm$1.16} & 6.94{ \scriptsize $\pm$0.42} & 90.17{ \scriptsize $\pm$0.75} & 2.46{ \scriptsize $\pm$0.13} & - & -  \\
    InfLoRA & 88.51{ \scriptsize $\pm$0.47} & 5.92{ \scriptsize $\pm$0.74} & 93.69{ \scriptsize $\pm$0.84} & 3.57{ \scriptsize $\pm$0.57} & - & - \\
    Ours & {\bf 91.91}{ \scriptsize $\pm$0.48} & {\bf 2.82}{ \scriptsize $\pm$1.42} & {\bf 95.54}{ \scriptsize $\pm$0.77} & {\bf 1.21}{ \scriptsize $\pm$0.21} & {\bf 70.90}{ \scriptsize $\pm$0.86} & {\bf 2.43}{ \scriptsize $\pm$0.67} \\
     \bottomrule
    \end{tabular}
\end{table*}

\begin{figure*}[t!]
    \centering
    \includegraphics[width=0.93\linewidth]{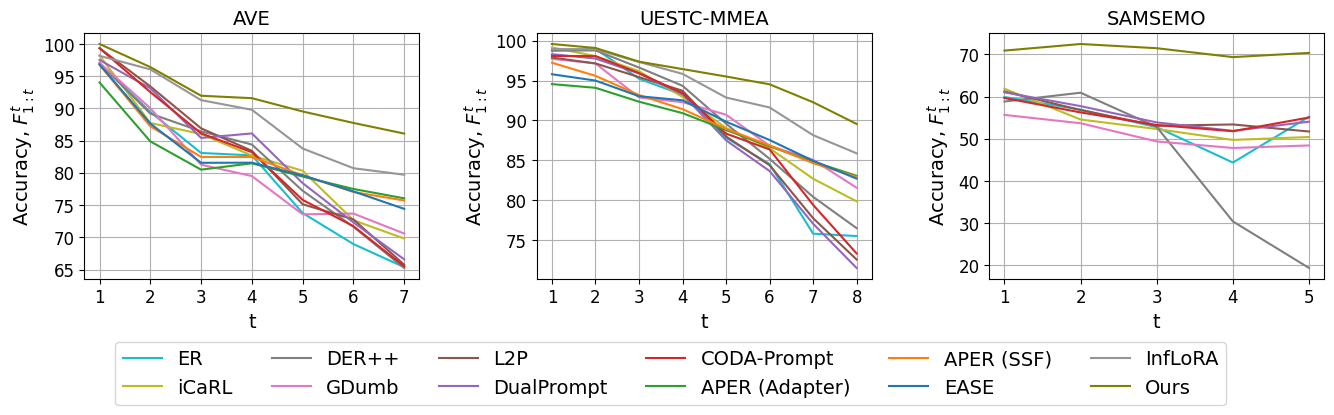}
    \caption{Accuracy on all seen tasks $F^t_{1:t}$ achieved by each method across time steps $t$ on the three datasets.}
    \label{fig:accuracy}
\end{figure*}

\paragraph{Evaluation Metrics.}
After each time step $t$, we evaluate the model on test samples of all tasks seen so far. We measure the classification accuracy using the weighted F1 score to account for class imbalance. 
Let $F_i^t$ denote the score for classifying samples of task $i$ using the model trained at step $t$ and $F_{i:j}^t$ be the score for classifying samples of tasks $i$ to $j$ using the model trained at step $t$.  
The overall accuracy is computed as 
\begin{equation}
  Acc=\frac{1}{T}\sum_{t=1}^{T}F_{1:t}^t  
\end{equation}
We quantify the amount of forgetting as the difference between the score achieved by the current model for each task and the maximum score obtained by any of the previous models on the corresponding task, that is, 
\begin{equation}
  Fgt=\frac{1}{T}\sum_{t=1}^{T}\Big[\frac{1}{t-1}\sum_{i=1}^{t-1}\max_{k\in[1,t-1]}(F_i^k-F_i^t)\Big].  
\end{equation}

\paragraph{Implementation Details.}
All experiments are conducted using PyTorch \cite{paszke2019pytorch} on a NVIDIA  H100 GPU.
We use the following PTMs for the respective modalities in each dataset: VideoMAE \cite{tong2022videomae} for the visual modality and AudioMAE \cite{huang2022masked} for the audio modality in AVE; VideoMAE \cite{tong2022videomae} for the visual modality and MonoSelfPAB \cite{logacjov2024self} for the inertial modality in UESTC-MMEA; and MARLIN \cite{cai2023marlin} for the facial visual modality, wav2vec \cite{baevski2020wav2vec} for the audio modality and multilingual BERT \cite{kenton2019bert} for the text modality in SAMSEMO.

All representations from the MMEncoder are projected to a dimension of 1024, and the fusion layers consist of a 2048-dimensional hidden layer followed by a 1024-dimensional output layer. 
Within each cross-adapter layer, we adopt 10 experts and activate the top-2 based on their highest contribution scores. We set the threshold for determining which experts to freeze at the end of each task to 10\% of the total samples in the current dataset.
We train the model using Adam optimizer, with a batch size of 12 for 20 epochs. The learning rate decays from 0.0001 with cosine annealing. 
Our hyperparameters $(\lambda_1,\lambda_2,\lambda_3)$ are set to $(1.0,1.0,10)$ for AVE and UESTC-MMEA, and $(0.1,1.0,5)$  for SAMESEMO.
Code is available at https://github.com/EvelynChee/MMEncoder.

\subsection{Comparative Study}

We compare our approach with traditional continual learning methods: ER \cite{chaudhry2019efficient}, iCaRL \cite{rebuffi2017icarl}, DER++ \cite{buzzega2020dark} and GDumb \cite{prabhu2020gdumb}.
We also compare our approach against recent PTM-based methods, including prompt-based approaches such as L2P \cite{wang2022learning}, DualPrompt \cite{wang2022dualprompt}, and CODA-Prompt \cite{smith2023coda}, as well as adapter-based methods like APER \cite{zhou2024revisiting}, EASE \cite{zhou2024expandable}, and InfLoRA \cite{liang2024inflora}.
To ensure a fair comparison, we allow these methods to use the same memory buffer for replaying data from previous tasks.
All methods use the same PTMs to initialize the respective modality encoders. 

Table \ref{tab:comparative_results} summarizes the performance of all methods across the three datasets. 
We see that our method achieves a significant improvement of 3.40\% and 1.85\% in $Acc$ over the next-best-performing baseline, InfLoRA, on the AVE and UESTC-MMEA datasets respectively.
It also achieves the lowest $Fgt$ values on both datasets, demonstrating strong knowledge retention.
The performance gains in class-incremental learning underscore the effectiveness of our method in consolidating and preserving knowledge across multiple modalities as new classes are introduced.

In domain-incremental learning on the SAMSEMO dataset, our method delivers a significant 15.18\% improvement in $Acc$ over the next-best-performing baseline, DualPrompt, while also achieving the lowest $Fgt$.
This highlights our model's ability to effectively adapt to new languages via modality interactions while maintaining robust performance on previously learned ones.

Figure \ref{fig:accuracy} shows the accuracy for each method across time steps, evaluated on the tasks learned up to that point.
Our approach achieves consistent improvement in classification performance across all time steps.
Notably, the gap between our approach and baseline methods widens as more classes are introduced in the AVE and UESTC-MMEA experiments.
Compared to the next-best-performing baseline InfLoRA, our method achieves a 6.38\% increase in final accuracy (from 79.72\% to 86.10\%) on the AVE dataset, and a 3.68\% increase (from 85.87\% to 89.55\%) on the UESTC-MMEA dataset.
For SAMSEMO, we also observe a significant 15.31\% improvement  in final accuracy (from 55.05\% to 70.36\%) over CODA-Prompt.

\begin{table}[t!]
\caption{Performance of modality adapters on AVE dataset. }
\label{tab:effectiveness_adapter}
    \centering
    \setlength\tabcolsep{2.5pt}
    \begin{tabular}{>{\raggedright\arraybackslash}m{4cm}|>{\centering\arraybackslash}m{1.6cm}>{\centering\arraybackslash}m{1.6cm}}
    \toprule
    & $Acc$ & $Fgt$ \\
    \midrule
    No adapter & 87.57{ \scriptsize $\pm$0.48} & 2.91{ \scriptsize $\pm$1.00} \\
    Modality-specific adapter & 88.39{ \scriptsize $\pm$1.03} & 2.93{ \scriptsize $\pm$0.48}\\
    Cross-modality adapter (Ours) & 91.91{ \scriptsize $\pm$0.48} & 2.82{ \scriptsize $\pm$1.42} \\
    \bottomrule
    \end{tabular}
\end{table}

\subsection{Effectiveness of Cross-Modality Adapter}

We next evaluate the effectiveness of our proposed cross-modality adapter. Table \ref{tab:effectiveness_adapter} shows the results on the AVE dataset. 
Our approach demonstrates a 4.34\% improvement in $Acc$ compared to a baseline that does not employ any adapters.
It also outperforms the modality-specific adapter introduced in \cite{yu2024boosting}, which uses independent adapters within the transformer layers for each modality, achieving a significant 3.52\% improvement in $Acc$. 
Our findings demonstrate the effectiveness of the cross-modality adapter in multi-modal continual learning, emphasizing the importance of learning the interaction between modalities and jointly adapting features, rather than treating each modality independently.

\paragraph{Sensitivity of cross-modality adapter to the number of experts.}
We assess the sensitivity of our approach by varying the number of experts in each cross-modality adapter. Table \ref{tab:sensitivity_experts} presents the results on the AVE dataset for different expert counts, ranging from 6 to 14. We observe that the model’s performance, both in terms of $Acc$ and $Fgt$, remains relatively stable across this range. 
This demonstrates the robustness of our approach to the number of experts used.

\begin{table}[t!]
\caption{Effect of number of experts on AVE dataset.}
\label{tab:sensitivity_experts}
    \centering
    \setlength\tabcolsep{4.2pt}
    \begin{tabular}{>{\centering\arraybackslash}m{2.5cm}|>{\centering\arraybackslash}m{2.2cm}>{\centering\arraybackslash}m{2.2cm}}
    \toprule
    \#Experts & $Acc$ & $Fgt$ \\
    \midrule
    6 & 91.39{ \scriptsize $\pm$0.80} & 2.91{ \scriptsize $\pm$0.87} \\
    8 & 91.24{ \scriptsize $\pm$0.66} & 3.11{ \scriptsize $\pm$1.34}  \\
    10 & 91.91{ \scriptsize $\pm$0.48} & 2.82{ \scriptsize $\pm$1.42} \\
    12 & 91.24{ \scriptsize $\pm$0.60} & 2.40{ \scriptsize $\pm$0.93}  \\
    14 & 91.32{ \scriptsize $\pm$0.88} & 2.48{ \scriptsize $\pm$0.68} \\
    \bottomrule
    \end{tabular}
\end{table}

\begin{table}[t!]
\caption{Performance comparison of the proposed representation alignment loss versus conventional contrastive loss on AVE dataset.}
\label{tab:effectiveness_align}
    \centering
    \renewcommand{\arraystretch}{1.1}  
    \begin{tabular}{>{\raggedright\arraybackslash}m{3.5cm}|>{\centering\arraybackslash}m{1.6cm}>{\centering\arraybackslash}m{1.6cm}}
    \toprule
    & $Acc$ & $Fgt$ \\
    \midrule
    Multi-modal supervised contrastive loss & 90.46{ \scriptsize $\pm$0.81}& 2.92{ \scriptsize $\pm$1.25}\\
    Representation alignment loss, $\mathcal{L}_{align}$ (Ours) & 91.91{ \scriptsize $\pm$0.48} & 2.82{ \scriptsize $\pm$1.42} \\
    \bottomrule
    \end{tabular}
\end{table}

\subsection{Effectiveness of Representation Alignment}
\label{sec:effectiveness_align}

We also evaluate the effectiveness of our proposed representation alignment loss, $\mathcal{L}_{align}$, by comparing it to the conventional multi-modal supervised contrastive loss. 
Table \ref{tab:effectiveness_align} shows the results on the AVE dataset. 
When the contrastive loss is used in place of $\mathcal{L}_{align}$, we observe a 1.45\% decrease in $Acc$. 
This performance gap underscores the effectiveness of $\mathcal{L}_{align}$ in promoting more robust multi-modal representations, enabling the model to better differentiate between previously learned and newly introduced classes.

\begin{figure}[t!]
    \centering
    \includegraphics[width=0.99\linewidth]{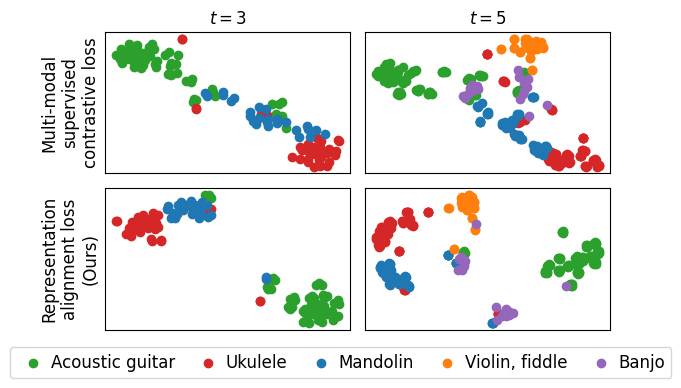}
    \caption{t-SNE visualization of joint representations learned using the conventional contrastive loss and our representation alignment loss.}
    \label{fig:rep_joint}
\end{figure}

This observation is supported by Figure~\ref{fig:rep_joint}, which visualizes the joint representations learned using the two losses. 
We focus on semantically similar classes, particularly string instruments, introduced at $t=3$ and $t=5$.
The model trained with $\mathcal{L}_{align}$ yields more compact and well-separated class clusters, showing improved discriminability even as learning progresses. 
In contrast, the conventional contrastive loss leads to greater confusion among the classes.

\begin{figure}[t!]
    \centering
    \includegraphics[width=0.85\linewidth]{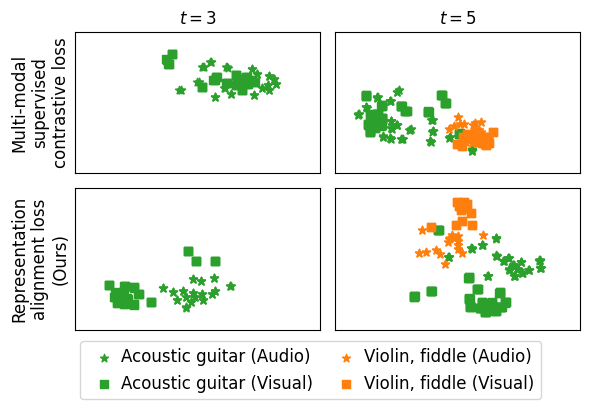}
    \caption{t-SNE visualization of modality-specific representations learned using the conventional contrastive loss and our representation alignment loss.}
    \label{fig:rep_mod}
\end{figure}

Figure~\ref{fig:rep_mod} further visualizes the modality-specific representations of two string instruments classes, Acoustic guitar and Violin, fiddle. 
In addition to class separability, $\mathcal{L}_{align}$ also results in separability between modalities of the same class, reinforcing the claim that it reduces loss of modality-specific information.

\subsection{Ablation Study on Loss Components}
Lastly, we conduct an ablation study to evaluate the impact of each loss component on the model performance by systematically excluding them from training. Table \ref{tab:ablation_loss} shows the results on the AVE dataset. 
We observe that removing the proposed $\mathcal{L}_{align}$ consistently leads to a reduction in $Acc$. On the other hand, excluding the preservation loss $\mathcal{L}_{preserve}$ results in an increase in $Fgt$. 
This highlights the complementary roles of our two losses, with $\mathcal{L}_{align}$ facilitating the acquisition of new knowledge by promoting the learning of robust multi-modal representations, while $\mathcal{L}_{preserve}$ helps retain previously learned information.
The best overall performance is achieved when both losses, along with $\mathcal{L}_{distil}$, are included.

\begin{table}[t!]
    \centering
    \caption{Effect of removing loss components on AVE dataset.}
    \renewcommand{\arraystretch}{1.1}  
    \setlength\tabcolsep{4.2pt}
    \begin{tabular}{>{\centering\arraybackslash}m{1.1cm}>{\centering\arraybackslash}m{1.35cm}>{\centering\arraybackslash}m{1.1cm}|>{\centering\arraybackslash}m{1.5cm}>{\centering\arraybackslash}m{1.5cm}}
    \toprule
    $\mathcal{L}_{align}$ & $\mathcal{L}_{preserve}$ & $\mathcal{L}_{distil}$ & $Acc$ & $Fgt$ \\
    \midrule
    \ding{51} & \ding{51} & \ding{51} & 91.91{ \scriptsize $\pm$0.48} & 2.82{ \scriptsize $\pm$1.42} \\ 
    \ding{51} & \ding{55} & \ding{51} & 90.79{ \scriptsize $\pm$1.20} & 3.98{ \scriptsize $\pm$1.39} \\
    \ding{55} & \ding{51} & \ding{51} & 88.85{ \scriptsize $\pm$0.79} & 2.80{ \scriptsize $\pm$0.71} \\
    \ding{55} & \ding{55} & \ding{51} & 89.66{ \scriptsize $\pm$0.32} & 3.18{ \scriptsize $\pm$0.66}  \\
    \ding{51} & \ding{51} & \ding{55} & 89.39{ \scriptsize $\pm$0.29}  & 6.77{ \scriptsize $\pm$1.04}  \\    
    \ding{51} & \ding{55} & \ding{55} & 88.88{ \scriptsize $\pm$1.10} & 7.36{ \scriptsize $\pm$0.95} \\
    \ding{55} & \ding{51} & \ding{55} & 87.54{ \scriptsize $\pm$0.46} & 6.82{ \scriptsize $\pm$0.57} \\   
    \ding{55} & \ding{55} & \ding{55} & 87.96{ \scriptsize $\pm$0.16}  & 7.12{ \scriptsize $\pm$0.29}  \\  
    \bottomrule
    \end{tabular}
    \label{tab:ablation_loss}
\end{table}

\section{Conclusion \& Future Work}

We present a framework for multi-modal continual learning that is built on pre-trained models and introduce cross-modality adapters to effectively capture interactions between modalities.
By employing a mixture-of-experts structure, the adapters facilitate continual adaptation to new tasks while preserving previously acquired knowledge. 
We design a representation alignment loss that minimizes information loss through a relaxed modality alignment constraint, while alleviating forgetting by maintaining consistency in the relationships between previously learned representations.
Extensive experiments across various continual learning scenarios and datasets involving diverse modalities show that our approach outperforms baseline methods in both improving overall accuracy and mitigating forgetting.
Future work will explore removing the use of memory buffers to improve scalability and broaden applicability.






\bibliography{ref}

\end{document}